\documentclass[fleqn,10pt]{olplainarticle}

\title{Evolving Complexity is Hard}

\author[1]{Alden H. Wright}
\author[2]{Cheyenne L. Laue}
\affil[1]{Department of Computer Science, University of Montana, Missoula, MT 59812, USA}
\affil[2]{Department of Computer Science, University of Montana, Missoula, MT 59812, USA}


\begin{abstract}
Understanding the evolution of complexity is an important topic in a wide variety of academic fields.
Implications of better understanding complexity include increased knowledge of major evolutionary transitions
and the properties of living and technological systems.  Genotype-phenotype (G-P) maps are fundamental to
evolution, and biologically-oriented G-P maps have been shown to have interesting and often-universal
properties that enable evolution by following phenotype-preserving walks in genotype space.   Here we
use a digital logic gate circuit G-P map where genotypes are represented by circuits and phenotypes by
the functions that the circuits compute.   We compare two mathematical definitions of circuit and phenotype
complexity and show how these definitions relate to other well-known properties of evolution
such as redundancy, robustness, and evolvability.  Using both Cartesian and Linear genetic programming
implementations, we demonstrate that the logic gate circuit shares many universal properties of biologically derived
G-P maps, with the exception of the relationship between one method of computing phenotypic evolvability,
robustness, and complexity. Due to the inherent structure of the G-P
map, including the predominance of rare phenotypes, large interconnected neutral networks, and the high
mutational load of low robustness, complex phenotypes are difficult to discover using evolution.
We suggest, based on this evidence, that evolving complexity is hard and we discuss computational
strategies for genetic-programming-based evolution to successfully find genotypes that map to complex
phenotypes in the search space.
\end{abstract}

\begin{document}

\flushbottom
\maketitle
\thispagestyle{empty}

\section{Introduction}

One of the important questions in the study of evolution is whether there is an inherent tendency for increased
complexity in living systems. Despite the intuitive idea that organic complexity defines the world around us, relatively simple bacteria and
archaea continue to be common and adaptive in many environments as well. The increase in complexity from prokaryotes to eukaryotes and from unicellular organisms
to multi-cellular organisms may have been achieved through a series of major evolutionary transitions as described in \cite{Szathmary1995a,Szathmary1995b}.

According to the arrow of complexity hypothesis defined by \cite{Bedau2009}, evolution works to increase the complexity of already complex organisms, indicating that complexity may increase in some lineages but not in others. How then, and why, does complexity emerge as an evolutionary strategy in some cases but not others?

Further issues with understanding the evolution of complexity involves defining and measuring precisely what is meant by the term.  Kolmogorov complexity is defined as the minimum length of a program that computes a specific function or string \cite{Li2008}. While this definition is foundational to work in a number of computational disciplines, it assesses maximum complexity to a completely random string. More recent approaches emphasize the idea that complex systems are neither completely regular nor entirely random.  For example, a random string of letters, from this perspective, is no more complex than a string of letters that periodically repeats. As work by \cite{Tononi1998} claims, it is functional integration in specialized systems that denotes true complexity – in the case of a string of letters, a readable piece of text is both intelligible and complex, whereas a string of completely reugluar or entirely random text is not. 

These ideas that intelligibility, information, and environment are inherently linked in complex systems, underscores the notion that the process of evolution is fundamentally based on the transfer of information through time.  In biological evolution, this information is, of course, stored in the genotypes of individuals, which map to phenotypes that in turn enable fitness-based survival and reproduction in specific environments. This correspondence, or mapping, referred to as the genotype-phenotype (G-P) map, enables phenotypes to adjust when the underlying genotypic information is altered and provides the basis for adaptive change. 

While evolutionary computation models tend to simplify phenotypes to fitness functions, we claim that the field can learn from biophysical models of G-P maps. Recent work demonstrates potentially universal structural properties of biophysical and biologically-related G-P maps (see \cite{Ahnert2017,Manrubia2021} for a review). This paper considers the structural properties of redundancy, robustness, and evolvability, which are by definition independent of selection and fitness in our models.  

Redundancy is defined in the literature as the number of genotypes that map to the same phenotype. 
High redundancy typically implies high robustness, which is defined as the ability to mutate a genotype without changing the phenotype that it maps to. 
Indeed, recent research on the structure of fitness landscapes demonstrates that those capable of realistically representing the evolution of biological processes and structures are extremely multidimensional and contain large neutral networks that connect high-fitness areas of the search space. 
This implies that evolution can find nearly any phenotype through neutral evolution alone, and without an associated “cost of selection” acquired through search space exploration of low-fitness areas, known as fitness valleys. 
Intuitively, both robustness and redundancy have significant implications for the property of evolvability, which implies the ability to evolve novel, adaptive phenotypes.  
The authors of \cite{Greenbury2021} show that for three biologically realistic genotype-phenotype map 
models—RNA secondary structure, protein tertiary structure and protein complexes- even with random fitness assignment, 
fitness maxima can be reached from almost any other phenotype without passing through fitness valleys.  

Despite the possibility of reaching any novel, adaptive form using neutral search, the reality is that complex phenotypes are both computationally and theoretically hard to evolve and a not insignificant question arises out of the research presented above - How does complexity evolve on highly redundant, neutral landscapes with corresponding high robustness, especially when highly complex phenotypes have inherently low redundancy and are both difficult to find using evolutionary search and easy to break using mutation?

In \cite{Wright2021} and in section \ref{subsubsection:evolvability}
we define two related measures of the complexity of genotypes and phenotypes for the logic gate 
circuit G-P map.  In comparing these definitions computationally, we found that complexity is strongly related to 
other structural properties of the G-P maps. Most importantly, complex phenotypes are rare, as they are represented 
by relatively few genotypes, and are thus hard to find relative to simpler genotypes in the search space. 
The phenotypic evolvability (defined in section 
\ref{subsubsection:evolvability}) must be approximated, which requires finding genotypes that map to the phenotype.  This can 
be done by either random sampling or by evolution.  If sampling is used, complexity is negatively related to 
phenotype evolvability, while if evolution is used, the relationship is positive.  We show that, particularly for Cartesian Genetic Programming (CGP), evolution-based 
exploration of pervasive neutral networks in the genotype space is a very effective strategy for the 
discovery of rare/complex phenotypes.

\section{The digital circuit G-P map}
\label{section:digital_circuit_gp_map}
Digital logic gate circuit G-P maps have been widely used to study the properties of evolution  \cite{Ofria2005,Arthur2006,Macia2009,Miller2009,Raman2011,Hu2012,Hu2018}.  
For the map used in this paper, genotypes are single-output feed-forward 
circuits of logic gates, such as AND and OR gates, and phenotypes are
the Boolean functions computed by circuits over all possible inputs to the circuit.  Phenotypes can be represented
as binary strings of length $2^n$ where $n$ is the number of inputs to the circuit.  Thus, there are $2^{2^n}$ phenotypes
for $n$-input $1$-output circuits.

Our logic-gate circuits use 5 logic gates:  AND, OR, NAND, NOR, and XOR.  One exception is Figure~\ref{fig:lredund_vs_rank4x1_CGP_LGP_noXOR} 
which uses only the first 4 of these gates.

Table \ref{tab:gates} shows the truth tables of these gates.  \begin{table}
  \centering
  \begin{tabular}{|c|c|c|c|c|c|c|}
    \hline
    \textit{X} & \textit{Y} & \textit{X} AND \textit{Y} & \textit{X} OR \textit{Y} & \textit{X} NAND \textit{Y} & \textit{X} NOR \textit{Y} & \textit{X} XOR \textit{Y} \\
    \hline
    1 & 1 & 1 & 1 & 0 & 0 & 0 \\
    \hline
    1 & 0 & 0 & 1 & 1 & 0 & 1 \\
    \hline
    0 & 1 & 0 & 1 & 1 & 0 & 1 \\
    \hline
    0 & 0 & 0 & 0 & 1 & 1 & 0 \\
    \hline
  \end{tabular}
  \caption{Truth tables of logic gates }
  \label{tab:gates}
\end{table}

\begin{figure}
  \centering
  \includegraphics[width=8cm]{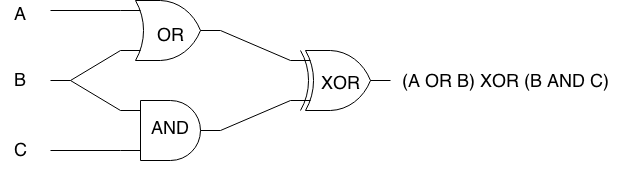}
  \caption{Example logic-gate circuit}
  \label{fig:circuit}
\end{figure}   

We compare results using two different genetic programming (GP) circuit representations: Cartesian (CGP) and Linear (LGP).  
The LGPvresults are new to this paper, while the CGP results are based on \cite{Wright2021} with some results
that are new to this paper.  

Our CGP representation is based on \cite{Miller2000} with one row of gates.  The levels-back parameter can be chosen
to be any integer from one (in which case, nodes can only connect to the previous layer) to the maximum number of nodes (in
which case a node can connect to any previous node) \cite{Miller2009}. Our LGP representation is described in \cite{Hu2020} with the exceptions
that we are using 10 instructions instead of 6, and we use the first 2 registers as computational registers and the remainder
as input registers.  Circuits are evaluated using bitwise instructions applied to unsigned integers.  

A CGP text representation of the circuit of Figure~\ref{fig:circuit} is: \texttt{circuit((1,2,3), ((4,OR,1,2), (5,AND,2,3), (6,XOR,4,5)))}.
A gate is represented as a 4-tuple composed of the gate number, the gate function, the first input, and the second input.
The last gate is the output gate.  Gates are evaluated by applying the gate function to the recursively evaluated inputs to the gate.
A circuit is evaluated by evaluating the output gate.  Circuits process all of the possible inputs to the circuit by using the bitwise
operations that are available on current computers.  Thus, the state of a gate of an $n$-input circuit is represented as a length $2^n$ 
bit vector which is stored as an unsigned integer.  Each input node of the circuit is initialized to an unsigned integer context, and 
these contexts are chosen so that the circuit is supplied with every possible combination of inputs.  For a 2-input circuit, these are \texttt{0xc, 0xa}, 
for a 3-input circuit they are \texttt{0xf0, 0xcc, 0xaa}, and for a 4-input circuit they are: \texttt{0xff00, 0xf0f0, 0xcccc, 0xaaaa}.
For an example, see subsection~\ref{subsubsection:tononi_complexity}

An LGP representation for the same circuit is: \texttt{[(2, 1, 3, 4), (1, 2, 4, 5), (5, 1, 1, 2)]}.  A gate is represented
as a 4-tuple where the elements of the tuple are the index of the gate function, the index of the output register,
the index of the first input, and the index of the second input.  Gates are evaluated sequentially.  Computational registers
are initialized to zeros and the input registers are initalized as described for the inputs of CGP circuits. 
Of course, the output phenotype for this example is the same as for CGP. 

\subsection{Neutral and epochal evolution}

Neutral evolution is an evolutionary strategy for exploring the space of genotypes that map to a given phenotype.  For each generation of the algorithm, 
the current genotype is randomly mutated with a point mutation which either changes the gate type or a connection between gates.  If the mutated genotype 
maps to the same phenotype, it becomes the current genotype.  Otherwise, the current genotype is unchanged.  Generations continue until some 
is reached.

We call this strategy, which is a variant of the (1+1) evolution strategy, \textbf{neutral evolution}.  This involves a random neutral walk 
in the network of the starting genotype and the associated phenotype.   We show below that for the digital circuit G-P map that CGP neutral evolution 
is highly effective at exploring genotype space.  This is consistent with results that show that neutral search is fundamental to the success of CGP  
\cite{Miller2006}.

Neutral evolution can be extended to an evolutionary strategy to find a genotype that maps to a given target phenotype.  In this case neutral evolution starts
from a random genotype.When the current genotype is mutated, we check to see if the mutated genotype maps to a phenotype which is Hamming distance closer 
to the target phenotype than the phenotype of the current genotype, and if so, we transition to neutral evolution starting with the mutated genotype. 
Otherwise, we continue neutral evolution based on the current genotype.  The algorithm terminates when a genotype mapping to the target genotype
is found or a step limit is reached.  Based on \cite{Crutchfield2002} we call this algorithm \textbf{epochal evolution}.  

These models, which do not use a population, are roughly equivalent to population-based evolution with SSWM (strong selection weak mutation) assumptions
where the population is almost always isogenic.  ``This is due to the fact that each new mutant will either fix, replacing the genotype shared by the whole population, 
or become extinct before another arises. Deleterious mutations fix with such low probability that this may be assumed never to happen" \cite[p.5]{Nichol2019}.

\cite{Miller2006} show that the most evolvable CGP representation is extremely large where over 95\% of gates are inactive.
Using CGP epochal evolution we are consistently able to evolve random 6-input 1-output phenotypes and the 7-input even parity problem by using 80 gates with 40 levels-back.

\section{Genotype-phenotype maps}
\vspace{2pt}

\subsection{Phenotype network and neutral sets}

The \textbf{phenotype network} of a G-P map has phenotypes as networks and single-point mutations as edges.
The \textbf{neutral set} of a phenotype is the set of genotypes that map to that phenotype.  The neutral set
may be path disconnected, although for the digital circuit G-P map with our parameter settings, it appears that
neutral sets are either connected or have one connected component that is much larger than other components.
\subsection{Redundancy}
All of the G-P maps studied in the review articles \cite{Ahnert2017} and \cite{Manrubia2021} have the fundamental property of redundancy which is a prerequisite for the additional properties
given below.  \textbf{Redundancy} is the property that there are many more genotypes than phenotypes.  Furthermore, the number of genotypes per phenotype varies widely
from phenotype to phenotype.  We show that for the CGP and LGP circuit G-P maps the frequency of genotypes per phenotypes can vary by at least 9 orders of magnitude for 
3-input and 4-input circuits.

\subsection{Robustness}
Robustness can be defined for both genotypes and phenotypes \cite{Wagner2008}.  The \textbf{robustness of a genotype} is defined as the fraction of mutations of the genotype 
that don't change the mapped-to phenotype.  The robustness of a phenotype is the average robustness of the genotypes that map to it.

Both \cite{Ahnert2017} and \cite{Manrubia2021} cite a linear relationship between the logarithm of the redundancy of a phenotype and its robustness as a 
universal property of G-P maps.  We confirm this relationship for the CGP and LGP circuit G-P maps (see Figure~\ref{fig:robustness_vs_lredund_vs_rank4x1_CGP_LGP}.)

\subsection{Evolvability} \label{subsubsection:evolvability}

A system is evolvable if mutations can produce adaptive and heritable phenotypic variation.  However, in the study of the structural properties of G-P maps,
it is hard to define what is adaptive because structural properties are independent of fitness.  Thus, a definition of evolvability for G-P maps 
should consider properties that are independent of selection.  Andreas Wagner \cite{Wagner2008} defines a system as evolvable
if mutations can produce heritable phenotypic variation.  More specially, he defines the \textbf{evolvability of a genotype} as the number of unique phenotypes produced
by mutation of the genotype.  This seems to imply an antagonistic relationship between robustness and evolvability---mutations producing many unique phenotypes
seems to imply a low likelihood of mutation preserving the phenotype.  And in fact, this is what is almost universally observed for genotype robustness and
evolvability in G-P map models.  

A way around this paradox is to look at \textbf{phenotype evolvability} which Wagner defines as the number of unique phenotypes in the mutational neighborhood of the given
neutral space of the phenotype.  This mutational neighborhood is found by mutating all of the genotypes that map to the given phenotype.  The evolvability of the phenotype
is the number of unique phenotypes produced by these mutations.  Both \cite{Ahnert2017} and \cite{Manrubia2021} find that phenotype evolvability and robustness
are positively related in the G-P maps that they study.  However, due to the vast size of genotype spaces, the evolvability of a phenotype cannot be computed exactly except in very trivial cases.  Thus, phenotype evolvability must be approximated, which necessitates finding genotypes that map to the given
phenotype.  Furthermore, \cite{Wright2021} shows a negative relationship between evolvability and robustness if epochal evolution is used to invert the G-P map in the approximation of phenotype evolvability.

\subsection{Universal structural properties}

\cite{Ahnert2017} reviewed structural properties for three biophysical G-P models: RNA secondary structure, the HP map
of protein folding, and the polynomial model.  The HP map simplifies protein folding by categorizing amino acids as either 
hydrophobic or hydrophillic.  The polynomial model \cite{Greenbury2014} is a two dimensional lattice model of self-assembly 
that can also be used as a G-P map to model protein quaternary structure.  

After an extensive study of many biophysical G-P maps and artificial life G-P maps, \cite{Manrubia2021}
proposed the following: 
``Some of the results highlighted in the former section hint at the possibility that any sensible G-P map (and, by
extension, artificial life system) is characterized by a generic set of structural properties that appear repeatedly, with
small quantitative variations, regardless the specifics of each map. Extensive research performed in recent years has
confirmed this possibility to an unexpected degree.

``Some of the commonalities documented are navigability, as reflected in the ubiquitous existence of large neutral
networks for common phenotypes that span the whole space of genotypes, a negative correlation between genotypic
evolvability and genotypic robustness, a positive correlation between phenotypic evolvability and phenotypic robustness, a linear 
growth of phenotypic robustness with the logarithm of the NSS [neutral set size], or a near lognormal distribution of the latter." 
\cite{Manrubia2021}  

We investigate these properties in the context of our CGP and LGP circuit models.  We confirm these properties with one
significant and surprising exception in regards to the relationship between robustness and evolvability which we explain below.

\section{Structural and complexity properties for the circuit G-P map}

\subsection{Redundancy and bias}

\begin{figure}
  \centering
  \includegraphics[width=12cm]{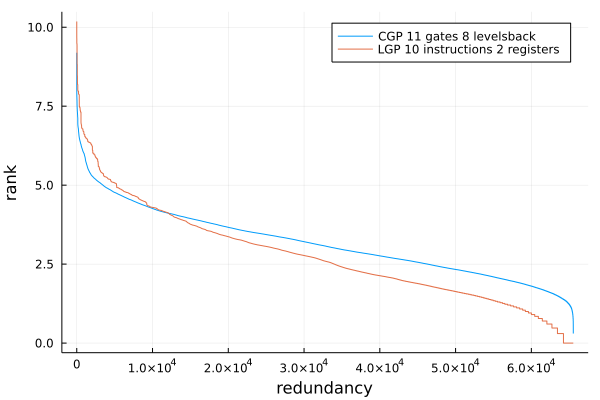}
  \caption{Log redundancy vs. rank by sampling with XOR gate. CGP: $10^{10}$ samples, LGP: $2 \times 10^{11}$ samples}
  \label{fig:lredund_vs_rank4x1_CGP_LGP}
\end{figure}

In this section we show the extreme variation in the frequency of digital circuit phenotypes.
Figure~\ref{fig:lredund_vs_rank4x1_CGP_LGP} shows the frequencies of all $2^{2^4} = 65536$
phenotypes ordered from most frequent to least frequent.  The CGP frequencies are based on
a sample of $10^{10}$ genotypes and the LGP frequencies are based on a sample of $2 \times 10^{11}$
genotypes.  Despite the much larger sample, rare phenotypes are much better represented using
CGP then using LGP.  There are 531 unrepresented phenotypes for LGP and none for CGP.

\begin{figure}
  \centering
  \includegraphics[width=12cm]{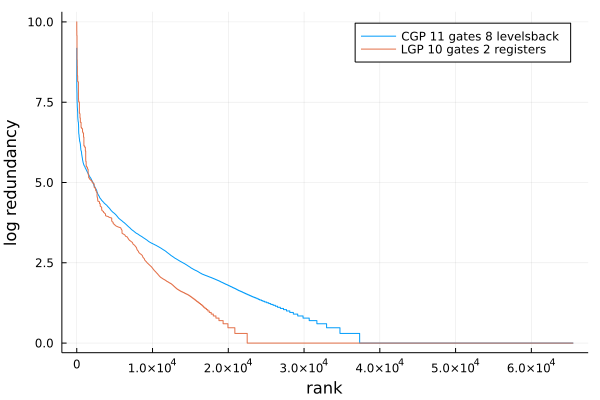}
  \caption{Log redundancy vs. rank sampling without XOR gate. CGP: $10^{10}$ samples, LGP: $2 \times 10^{11}$ samples}
  \label{fig:lredund_vs_rank4x1_CGP_LGP_noXOR}
\end{figure}

The choice of gate functions contributes to the number of phenotypes discovered by sampling.
While all of the other plots in this paper are generated using the 5 gate functions AND, OR, NAND, NOR, and XOR,
Figure~\ref{fig:lredund_vs_rank4x1_CGP_LGP_noXOR} uses only the first 4 of these gates.  In this case only about 1/3 of 
phenotypes for LGP and 2/3 for CGP despite the 20 times larger sample for LGP.  

\subsection{Robustness}
\begin{figure}
  \centering
  \includegraphics[width=12cm]{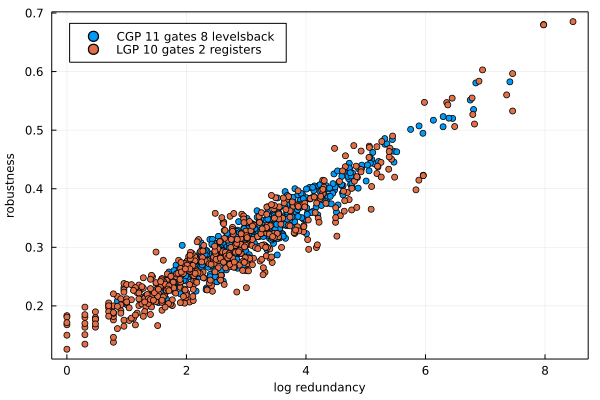}
  \caption{Robustness vs. log redundancy CGP 11 gates 8 lb LGP 10 instructions}
  \label{fig:robustness_vs_lredund_vs_rank4x1_CGP_LGP}
\end{figure}

Figure~\ref{fig:robustness_vs_lredund_vs_rank4x1_CGP_LGP} confirms the strong linear 
relationship between robustness and log redundancy which is universal in the G-P maps
reviewed by \cite{Ahnert2017}.

\subsection{Evolvability}
We approximated the evolvability of a phenotype by counting the total number of unique
phenotypes in the mutational neighborhoods of 600 genotypes.
These genotypes were found by using epochal evolution with the phenotype as the target.\cite{Wright2021} refers to this method as \textbf{evolution evolvability}.  
An alternative method, sampling evolvability, is based on using a sampling methodology
to find genotypes that map to the given phenotype is described in \cite{Wright2021}.
\cite{Hu2020} uses random walks to implement a sampling methodology.
As shown in Figure 7 of \cite{Wright2021}, evolution evolvability is 
strongly negatively related to log robustness while sampling evolvability is strongly positively 
related to log robustness.  The negative relationship between log redundancy and evolvability is different from the biologically related G-P maps described in
\cite{Ahnert2017,Manrubia2021}.

\begin{figure}
  \centering
  \includegraphics[width=12cm]{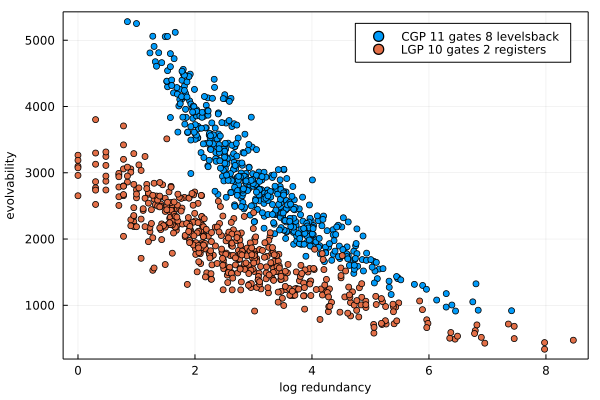}
  \caption{Evolution  evolvability vs Log redundancy CGP 11 gates 8 lb LGP 10 instructions}
  \label{fig:evolvability_log_redund_CGP_LGP}
\end{figure}

Figure~\ref{fig:evolvability_log_redund_CGP_LGP} shows a strong negative relationship
between log redundancy and phenotype evolvability for 500 random 4-input phenotypes.  The same
random phenotypes were used for CGP and LGP.  These results show that evolution 
evolvability is much less LGP than for CGP.

\begin{figure}
  \centering
  \includegraphics[width=12cm]{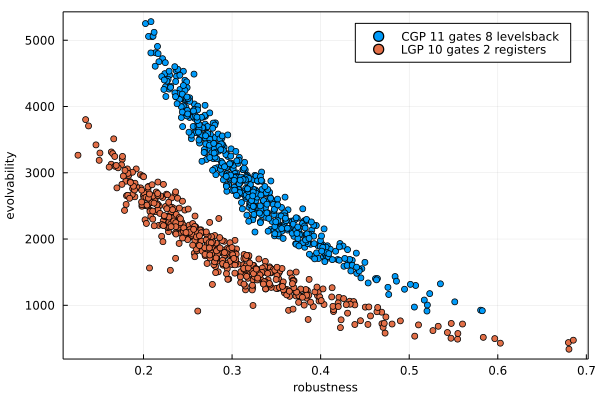}
  \caption{Evolution evolvability vs robustness CGP 11 gates 8 lb LGP 10 instructions}
  \label{fig:evolvability_robustness_CGP_LGP}
\end{figure}

Figure~\ref{fig:evolvability_robustness_CGP_LGP} shows a strong negative relationship between
evolution phenotype evolvability and robustness.  Again, LGP evolvability is much less than CGP evolvability.

\subsection{Tononi and Kolmogorov Complexity}
\label{subsection:complexity}

We use two methods to compute the complexity of the digital circuits representing phenotypes in our model. First, 
we apply the Tononi measure of complexity defined for neural circuits in \cite{Tononi1998}.

\subsubsection{Entropy and Mutual Information}

Based on work by \cite{Tononi1994,Tononi1998} the entropy $H(X)$ of a discrete random variable $X$ with $N$ states is given by
$$
H(X) = -\sum_{i=1}^N p_i \log(p_i)
$$
where $p_i$ is the probability of state $i$.   The entropy of $X$ is highest when all states of $X$ are distinct
and have equal probability.  It is lowest when $X$ has only one state.

The mutual information of two subsets $A$ and $B$ of a system $X$ is given by
$$
MI(A;B) = H(A)+H(B)-H(X)
$$
If $A$ and  $B$ are independent, then $MI(A;B)=0$, while if $A$ can be completely predicted knowing $B$, then $MI(A;B)= H(A)=H(B)=H(X)$.
Mutual information has the advantages that it is multivariate and captures both linear and non-linear dependencies.     

\subsubsection{Tononi Complexity}
\label{subsubsection:tononi_complexity}

Given a circuit, let $X$ be the binary matrix with rows corresponding to gates and columns corresponding to the $2^n$ possible inputs to the circuit.  
As described in section~\ref{section:digital_circuit_gp_map} the inputs to an $n$-input circuit are specified by the $n$ contexts which are the same
for any $n$-input circuit.
Each row of $X$ is the state of the corresponding gate when the circuit is executed.  

Table~\ref{tbl:truth_tables} is the complete state of the circuit of Figure~\ref{fig:circuit}
The first 3 rows correspond to the 3 standard contexts for a 3-input circuit, and the
matrix $X$ is the last 3 rows.  The hexadecimal representation of each row is given as the last column. 
The last row represents the output phenotype of the circuit.

\begin{table}
\begin{center}
\begin{tabular}
{|l|c c c c c c c c| l |}
\hline
Input 1 & 1 & 1 & 1 & 1 & 0 & 0 & 0 & 0 & \texttt{0xf0}\\
Input 2 & 1 & 1 & 0 & 0 & 1 & 1 & 0 & 0 & \texttt{0xcc}\\
Input 3 & 1 & 0 & 1 & 0 & 1 & 0 & 1 & 0 & \texttt{0xaa}\\
\hline
Gate 4 OR & 1 & 1 & 1 & 1 & 1 & 1 & 0 & 0 & \texttt{0xfc}\\
Gate 5 AND &1 & 0 & 0 & 0 & 1 & 0 & 0 & 0 & \texttt{0x88}\\
Gate 6 XOR &0 & 1 & 1 & 1 & 0 & 1 & 0 & 0 & \texttt{0x74}\\
\hline
\end{tabular}  
\caption{Truth tables of gates}
\label{tbl:truth_tables}
\end{center}
\end{table}

We compute the entropy of $X$ by interpreting $X$ as a probability distribution over the columns of $X$.  Each column is a binary vector which can
be interpreted as a bit string.  For the above example, there are two occurrences of \texttt{110}, four occurrences of \texttt{101}, and
two occurrences of \texttt{000}.   Thus, the probability of \texttt{110} is $1/4$, the probability of \texttt{101} is $1/2$, and the 
probability of \texttt{000} is $1/4$.   We find that the entropy $H(X)$ is given by
$H(X) = -1/4 \log(1/4) -1/2 \log(1/2) -1/4 \log(1/4) = 3/2$.  (We use base $2$ logarithms.)

Following \cite{Tononi1994,Tononi1998}, we will use $X_j^k$ to denote a sub-matrix of $X$ of cardinality $k$ where $j$ is an index over different subsets of cardinality $k$.  Continuing our example, let $X\{1,3\}$ denote the sub-matrix of $X$ corresponding to rows $1$ and $3$ of $X$.  Thus, $X\{1,3\}$ would be $X_j^2$ for some $j$.  We denote the entropy of $X\{1,3\}$ by $H\{1,3\}$ 
\begin{table}
\centering
X\{1,3\} = 
\begin{tabular}
{|c c c c c c c c|}
\hline
1 & 1 & 1 & 1 & 1 & 1 & 0 & 0 \\
0 & 1 & 1 & 1 & 0 & 1 & 0 & 0 \\
\hline
\end{tabular}           
\caption{Sub-matrix of $X$ corresponding to rows 1 and 3}
\end{table}

This matrix has columns \texttt{10}, \texttt{11}, and \texttt{00} with probabilities $1/4$, $1/2$, and $1/4$ respectively.
Thus, $H(A) = H\{1,3\} = 3/2$.  

If $M$ is the number of gates, there are $\binom{M}{k}=\frac{M!}{k!(M-k)!}$  ways of choosing $X_j^k$.

Tononi’s definition \cite{Tononi1994,Tononi1998} of the complexity $C(X)$ of a circuit whose matrix is $X$ is given by
\begin{equation} \label{eqn:complexity}
C(X)=\sum_{k=1}^{M/2} <MI(X_j^k;X-X_j^k)>\; =\frac{1}{2} \sum_{k=1}^{M} \frac{1}{\binom{M}{k}} \sum_{j=1}^{\binom{M}{k}} MI(X_j^k;X-X_j^k)
\end{equation}
where $< >$ denotes ensemble average (the average over $j$).
The papers \cite{Tononi1994,Tononi1998} give alternative equivalent information theoretic formulas for complexity.

To continue the example, the complexity of the circuit of Figure~\ref{fig:circuit} is computed as:

\begin{align*}
C(X)  & = & & MI(X\{1\};X\{2,3\}) + MI(X\{2\};X\{1,3\}) + MI(X\{3\};X\{1,2\}) \\
&= & & 1/3 ((H\{1\}+H\{2,3\}-H\{1,2,3\}) + (H\{2\}+H\{1,3\}-H\{1,2,3\})\\ 
&  & & + (H\{3\}+H\{2,3\}-H\{1,2,3\})\\
&= & & 1/3( 0.8113 + 0.8113 + 1.0) \\
&= & & 0.8742
\end{align*} 

The Tononi complexity of a 
digital circuit phenotype is the average of the Tononi complexity of a sample of circuits that map to the phenotype.  

\subsubsection{Kolmogorov complexity}

The second method we use to compute complexity is a version of Kolmogorov complexity designed for digital circuits.  The Kolmogorov complexity of a string is 
defined as the minimum length of a computer program that generates the string \cite{Li1989}. 

Digital circuit
phenotypes can be expressed as bit strings, and computer circuits are a language for generating these strings.  Based on this logic, we define the Kolmogorov complexity of a digital circuit phenotype as the minimum number of logic gates needed in a circuit to
map to a specified phenotype.  This assumes a specific representation (CGP or LGP) with a fixed 

levels-back or number of registers.  
As a simple example, there is no single gate CGP circuit that computes the function for the phenotype A EQV B (bit string 1001 = hex 0x9). However, since the CGP circuit ((1,2), ((3, XOR, 2,1), (4,NOR,3,3))) 
computes this phenotype, there is a 2-gate circuit that also computes the function.  Here we see that the Kolmogorov complexity 
of this phenotype is 2.  

\begin{figure}
  \centering
  \includegraphics[width=12cm]{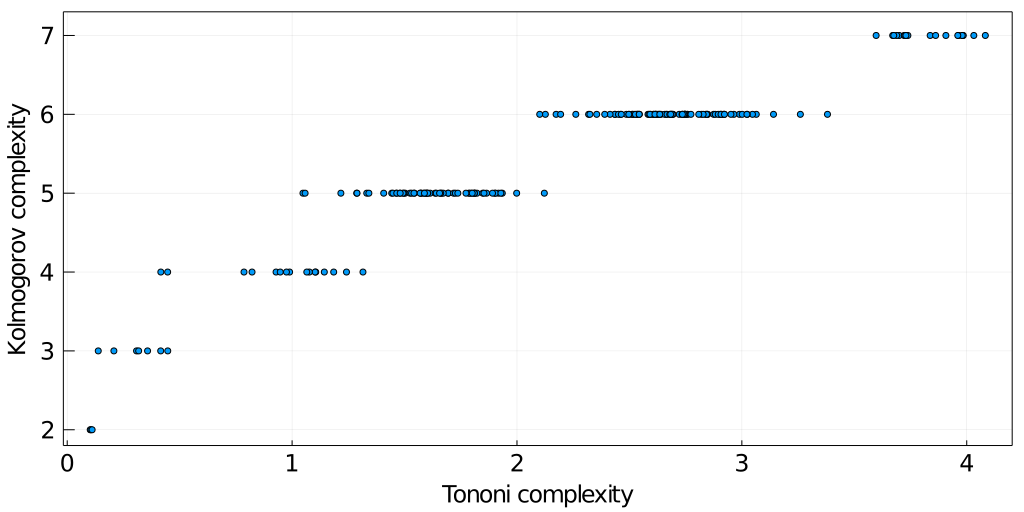}
  \caption{The relationship between Tononi and Kolmogorov complexity CGP 4 inputs 11 gates 8 levels-back}
  \label{fig:Tononi_Kolmogorov_complexity_cgp}
\end{figure}

Figure~\ref{fig:Tononi_Kolmogorov_complexity_cgp} shows that Tononi and Kolmogorov complexity are empirically consistent
for one setting of CGP parameters.  This plot is taken from Figure~3 of \cite{Wright2021}.

\cite{ahnert2010} define a complexity for the polynomial G-P map which is based on Kolmogorov complexity.

\cite{Dingle2018} show that evolving genotypes to phenotypes of high Kolmogorov complexity is computationally hard.
The authors consider computable maps of the form $f:I \rightarrow O$ where $I$ is a collection of input sequences and $O$ is the corresponding
collection of outputs which can be described as discrete sequences.  When applied to our definition of Kolmogorov complexity for
digital circuits, $I$ is the standard context for the number of inputs, and $f$ is the program that evaluates the circuit as a function of the context.
This program is part of our simulation code and does not change as the size of the input sequences increases.

If $x \in O$, Dingle et al. defined $P(x)$ to be the probability that a random input from $I$ will give will map to $x$.  Their main result is their equation (3):
$$
P(x) \leq 2^{-a \tilde{K}(x)-b}
$$
where $\tilde{K}$ is an approximate Kolmogorov complexity and the constants $a \geq 0$ and $b$ depend on the mapping $f$ and not on $x$.
The result is based on the assumption that $f$ is of limited complexity, which is demonstrated for digital circuits as the code for $f$
does not change as the input size increases.

It appears that the Dingle et al. result applies to digital circuit G-P maps in which case evolving complexity is hard from a computational
complexity point of view in a way that scales to all circuit input sizes.  Our simulation results confirm this theoretical prediction.
An application of the \cite{Dingle2018} theoretical result to G-P maps is \cite{Johnston2022}.  

\subsection{Complexity Density}

\begin{figure}
  \centering
  \includegraphics[width=12cm]{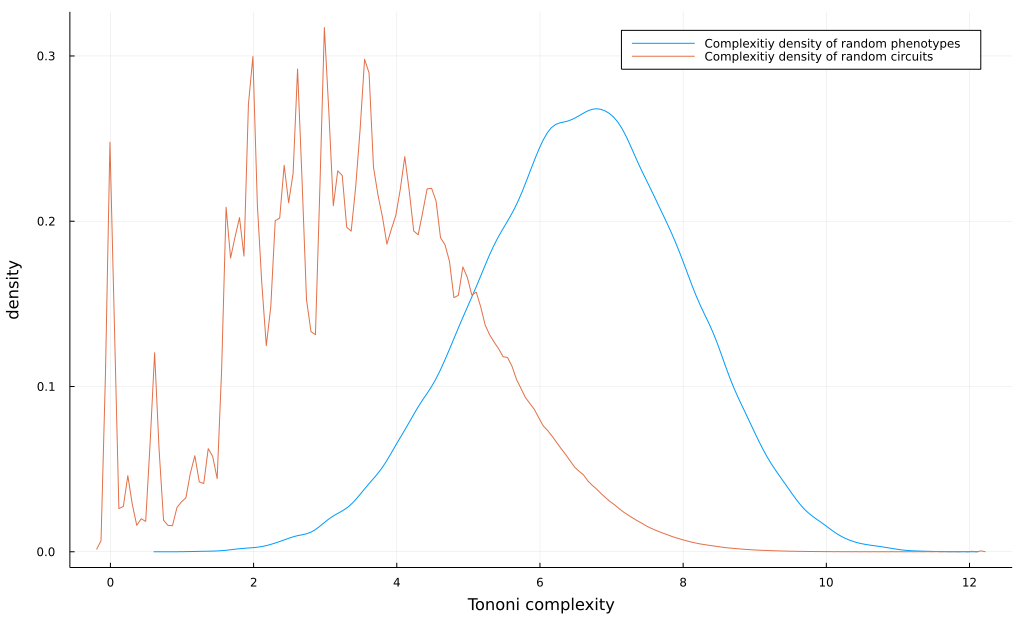}
  \caption{Complexity densities of genotypes and phenotypes CGP 4 inputs 11 gates 8 levels-back}
  \label{fig:t_complexity_density_cgp}
\end{figure}

\begin{figure}
  \centering
  \includegraphics[width=12cm]{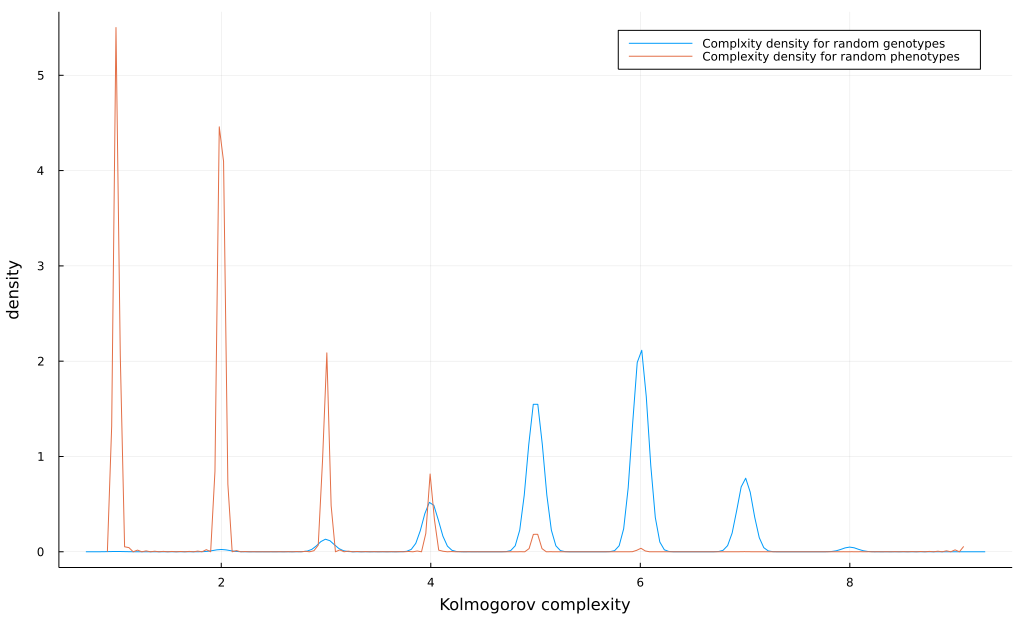}
  \caption{Kolmogorov complexity densities of genotypes and phenotypes CGP 4 inputs 11 gates 8 levels-back}
  \label{fig:k_complexity_density_cgp}
\end{figure}

Figures~\ref{fig:t_complexity_density_cgp} and \ref{fig:k_complexity_density_cgp} show that the Tononi and Kolmogorov complexity of random CGP genotypes is much less than the complexity of 
random phenotypes.  The Tononi complexity of phenotypes is computed by evolving genotypes that map to a phenotype, which may give an increased estimate of complexity in comparison to sampling.   Figure~\ref{fig:t_complexity_density_cgp} is taken from Figure 5 
of \cite{Wright2021}.

\subsection{Complexity Correlations}

\begin{figure}
  \centering
  \includegraphics[width=12cm]{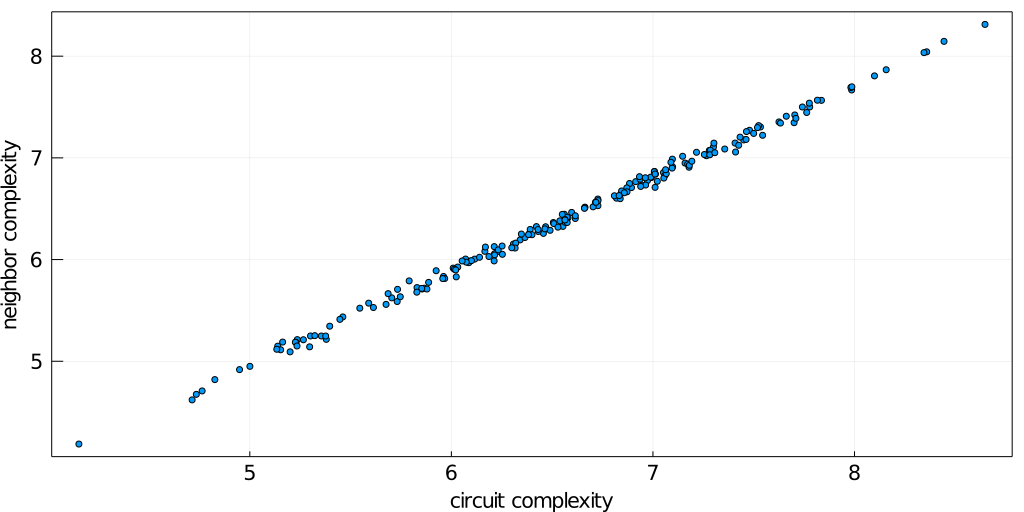}
  \caption{Average complexity of neighboring genotypes CGP 4 inputs 11 gates 8 levels-back}
  \label{fig:complexity_neighboring_genotypes_cgp}
\end{figure}

Figure~\ref{fig:complexity_neighboring_genotypes_cgp} shows that the average Tononi complexity of 1-mutant neighbors of a genotype
is close to the Tononi complexity of the genotype.  This figure is taken from Figure~4 of \cite{Wright2021}

\subsection{Complexity and redundancy}
\subsubsection{Complex phenotypes are rare}

\begin{figure}
  \centering
  \includegraphics[width=12cm]{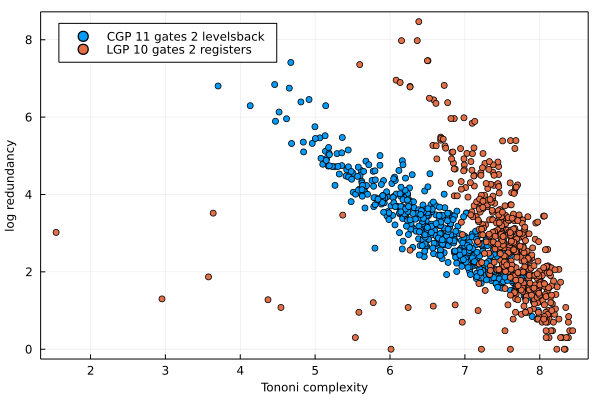}
  \caption{Log redundancy vs. complexity CGP 11 gates 8 lb LGP 10 instructions}
  \label{fig:lredund_vs_complexityCGP_LGP}
\end{figure}

Figure~\ref{fig:lredund_vs_complexityCGP_LGP} shows a strong negative relationship between Tononi complexity and log redundancy.
Note that the computation of Tononi complexity for CGP differs from that of LGP which is an explanation for the larger values of Tononi complexity for LGP. The same 500 random phenotypes were used for these plots as for the evolvability plots.

For both representations, complex phenotypes are rare, i. e., they are represented by a small number of genotypes, indicating that
they are hard to evolve.

\subsection{Complexity and robustness}
\subsubsection{Complex phenotypes have low robustness}

\begin{figure}
  \centering
  \includegraphics[width=12cm]{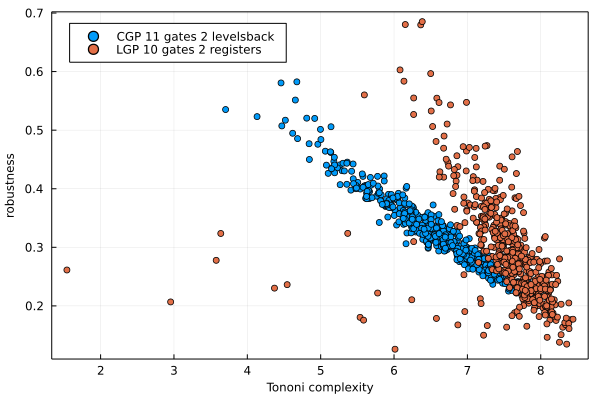}
  \caption{Robustness vs. complexity CGP 11 gates 8 lb LGP 10 instructions}
  \label{fig:robust_vs_complexityCGP_LGP}
\end{figure}

Figure~\ref{fig:robust_vs_complexityCGP_LGP} shows the strong negative relationship between robustness and Tononi complexity.
The plot is very similar to Figure~\ref{fig:lredund_vs_complexityCGP_LGP} which is not surprising given the strong positive relationship
between log redundancy and robustness shown in Figure~\ref{fig:robustness_vs_lredund_vs_rank4x1_CGP_LGP}.

\subsection{Complexity and evolvability}
\begin{figure}
  \centering
  \includegraphics[width=12cm]{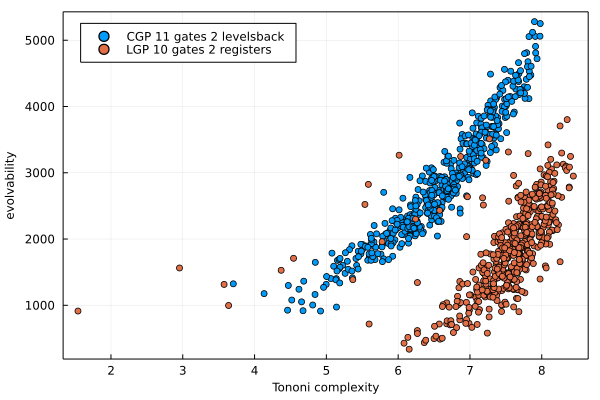}
  \caption{Evolution evolvability vs. complexity CGP 11 gates 8 lb LGP 10 instructions}
  \label{fig:evolvability_vs_complexityCGP_LGP}
\end{figure}

Figure~\ref{fig:evolvability_vs_complexityCGP_LGP} shows that evolution evolvability increases with Tononi complexity for both CGP and LGP. 
This confirms the result of Figure 8a of \cite{Wright2021}.  Note that sampling evolvability has a negative relationship 
with complexity as shown in Figure 8b of \cite{Wright2021}.

The positive relationship between evolution evolvability and complexity suggests that as complexity evolves, a greater number
of new phenotypes are nearby, and thus there is a positive feedback enabling the discovery of further complexity.

\section{Conclusion}

The review article \cite{Manrubia2021} hypothesizes that the evolutionary process is often more determined by the structure of 
the relevant G-P map than by selection.  Two examples are the ``survival of the flattest" phenomenon where genotypes with high robustness
(i. e., low mutational load) will out-compete genotypes with higher fitness but lower robustness. \cite{Wilke2001}, and the ``arrival of the frequent"
phenomenon where infrequent (rare) phenotypes are not discovered in time to compete with frequent (common) phenotypes \cite{Schaper2014}.
Thus, understanding the structural properties of G-P maps is important for understanding evolution.  

We have compared CGP and LGP representations of digital logic gate circuit G-P maps,
where genotypes are circuits and phenotypes are the functions computed by circuits.
We have shown that both representations have nearly all of the structural properties proposed as universal for biologically related
and inspired G-P maps. Our models show that redundancy varies by many orders of magnitude between phenotypes and in CGP models, even those containing rare (low redundancy) phenotypes, neutral sets tend to percolate through genotype space, and this is true although to a lesser
extent) for LGP.  Thus, exploring neutral sets is an effective strategy for evolving genotypes (circuits) that map to a given phenotype.

Neutral evolution is enabled by robustness which is the fraction of mutations of a genotype which do not change the mapped-to phenotype.
The evolvability of a phenotype is the number of unique phenotypes which are one mutation away from a genotype in the neutral set
of the phenotype.  We found a negative relationship between evolvability approximated using epochal evolution and robustness, 
which differs from results found in biologically-related G-P maps.  On the other hand, if evolvability is approximated by using
a large random sample of genotypes we show this relationship changes from 
negative to positive (which is the ``universal" result for biologically related G-P maps).  This was shown for CGP in our previous paper 
\cite{Wright2021}

We defined an information-theoretic measure of circuit (genotype) complexity based on \cite{Tononi1998} that we call Tononi complexity,
and we defined Kolmogorov complexity for phenotypes based on the minimum number of gates needed to compute a phenotype. 
Furthermore, we empirically demonstrated that these two measures of complexity are related.  

Why then is evolving genotypes that map to complex phenotypes so difficult?  First, random genotypes have low
complexity as shown in Figures~\ref{fig:t_complexity_density_cgp} and \ref{fig:k_complexity_density_cgp}.
Second, complex phenotypes tend to have both low redundancy and low robustness as shown in Figures 
\ref{fig:lredund_vs_complexityCGP_LGP} and \ref{fig:robust_vs_complexityCGP_LGP}.
Third, the computational complexity results of \cite{Dingle2018} show that the probability of sampling genotypes that map to a
phenotype decreases exponentially with the Kolomogorov complexity of the phenotype.

How then does complexity evolve at all? Our results indicate some strategies that facilitate the evolution of complex phenotypes.  

To begin, the average complexity of the genotypes
near to a given phenotype tends to be close to the complexity of a given genotype as shown in Figure~\ref{fig:complexity_neighboring_genotypes_cgp}.
Based on this, conducting evolutionary searches near complex phenotypes should lead to genotypes of high complexity as well.  
Second, our results demonstrating a positive relationship between complexity and evolution evolvability: see Figure~\ref{fig:evolvability_vs_complexityCGP_LGP}.
suggest that as genotypes evolve those that map to neutral sets
will be mutationally accessible from the genotypes of many different phenotypes.  

\subsection{Further work}

\begin{enumerate}
\item Determine whether the circuit G-P maps have the shape-space covering property.  A G-P map has the shape space covering property 
that, if given a phenotype, only a small radius around a genotype encoding that phenotype needs to be explored in order to find the most 
common phenotypes.  
\item Find additional computational problems whose G-P map has the universal structural properties described above.
\item Investigate the reason that genotypes evolved to map to a target phenotype have higher Tononi complexity than genotypes sampled to map to the target phenotype.
\end{enumerate}

\section{Acknowledgement*}
The input of reviewers Stu Card and Ting Hu was very helpful.  Cooper Craig produced some of the plots.  And we thank Jesse Johnson for the use of computational resources.



\bibliography{gptp_arXiv}

\end{document}